\PassOptionsToPackage{table}{xcolor}
\documentclass[sigconf]{acmart}
\AtBeginDocument{%
  }

\setcopyright{acmlicensed}
\copyrightyear{2026}
\acmYear{2026}
\acmDOI{XXXXXXX.XXXXXXX}
\acmConference[SIGIR '26]{Proceedings of the 49th International ACM SIGIR Conference on Research and Development in Information Retrieval}{July 20--24, 2026}{Melbourne | Naarm, Australia}
\acmBooktitle{Proceedings of the 49th International ACM SIGIR Conference on Research and Development in Information Retrieval (SIGIR '26), July 20--24, 2026, Melbourne | Naarm, Australia}
\acmISBN{978-1-4503-XXXX-X/2026/06}

\usepackage{multirow}
\usepackage{array}
\usepackage{arydshln}
\usepackage[most]{tcolorbox}
\usepackage[table]{xcolor}
\usepackage{enumitem}
\newcommand{\mymodel}{SmartSearch}

\begin{document}

\title{SmartSearch: Process Reward-Guided Query Refinement for Search Agents}

\author{Tongyu Wen}
\email{wentongyu@ruc.edu.cn}
\affiliation{%
  \institution{Renmin University of China}
  \city{Beijing}
  \country{China}
}

\author{Guanting Dong}
\email{dongguanting@ruc.edu.cn}
\affiliation{%
  \institution{Renmin University of China}
  \city{Beijing}
  \country{China}
}

\author{Zhicheng Dou}
\email{dou@ruc.edu.cn}
\affiliation{%
  \institution{Renmin University of China}
  \city{Beijing}
  \country{China}
}

\renewcommand{\shortauthors}{Trovato et al.}

\begin{abstract}
  Large language model (LLM)-based search agents have proven promising for addressing knowledge-intensive problems by incorporating information retrieval capabilities. Existing works largely focus on optimizing the reasoning paradigms of search agents, yet the quality of intermediate search queries during reasoning remains overlooked. As a result, the generated queries often remain inaccurate, leading to unexpected retrieval results and ultimately limiting search agents' overall effectiveness. To mitigate this issue, we introduce \textbf{\mymodel{}}, a framework built upon two key mechanisms: (1) \textbf{Process rewards}, which provide fine-grained supervision for the quality of each intermediate search query through Dual-Level Credit Assessment. (2) \textbf{Query refinement}, which promotes the optimization of query generation by selectively refining low-quality search queries and regenerating subsequent search rounds based on these refinements. To enable the search agent to progressively internalize the ability to improve query quality under the guidance of process rewards, we design a three-stage curriculum learning framework. This framework guides the agent through a progression from imitation, to alignment, and ultimately to generalization. Experimental results show that \mymodel{} consistently surpasses existing baselines, and additional quantitative analyses further confirm its significant gains in both search efficiency and query quality. The code is available at \href{https://github.com/MYVAE/SmartSearch}{https://github.com/MYVAE/SmartSearch}.
\end{abstract}




\maketitle

\section{Introduction}

Large language models (LLMs) have shown strong performance across a variety of tasks~\cite{achiam2023gpt,touvron2023llama,touvron2023llama2,brown2020language,chowdhery2023palm}, including translation~\cite{zhang2023prompting,xu2023paradigm}, summarization~\cite{zhang2024benchmarking,tang2023evaluating}, and question answering~\cite{singhal2023towards,kamalloo2023evaluating}. However, challenges remain, particularly with issues like hallucinations~\cite{ji2023survey,rawte2023survey} and the absence of recent or field-specific knowledge, which may result in inaccurate or outdated answers. Retrieval-augmented generation (RAG)~\cite{gao2023retrieval,chen2024benchmarking,jiang2025deepretrieval} has been introduced to address these challenges by incorporating external knowledge to complement the model's internal knowledge~\cite{schick2023toolformer,DBLP:journals/corr/abs-2505-06311}. However, static RAG faces limitations in its ability to handle more complex, dynamic, and deep exploration tasks.

\begin{figure}[t]
  \centering
  \includegraphics[width=\linewidth]{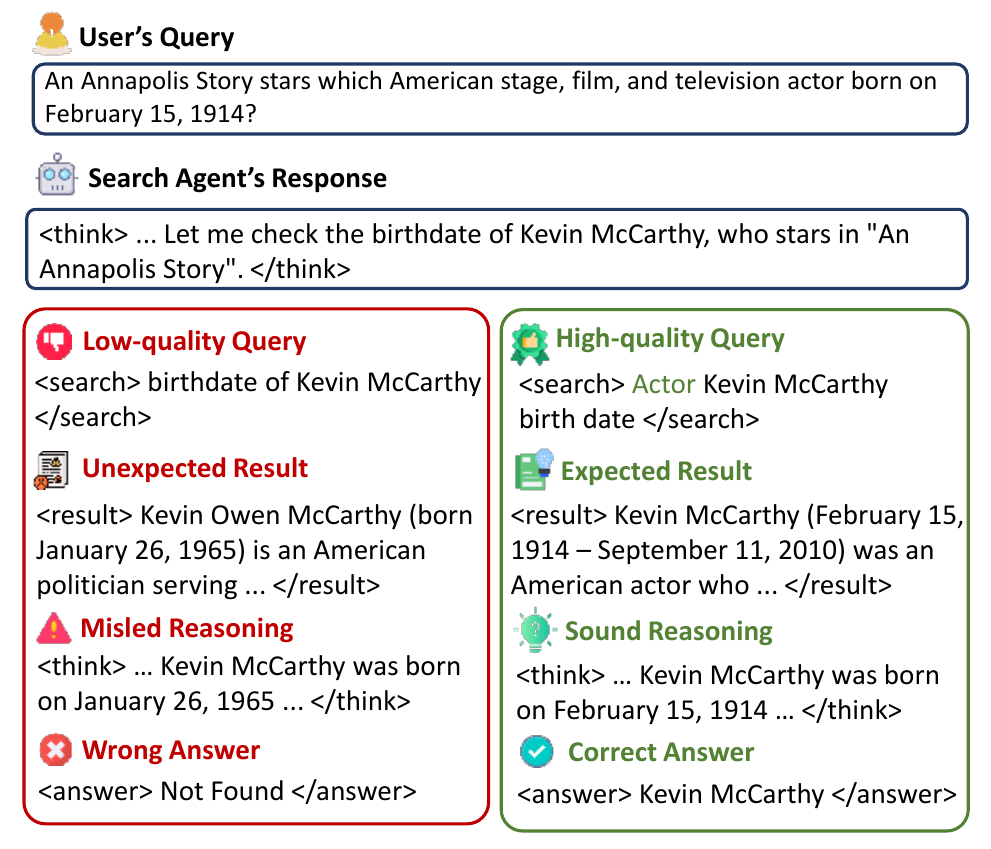}
  \caption{An example from ASearcher \cite{gaobeyond} dataset demonstrating how low-quality intermediate search queries lead to unexpected retrieval results and derail the entire trajectory.}
  \label{fig:overall}
\end{figure}

Recently, LLM-based search agents have proven to be a promising method \cite{li2025search,jinsearch,song2025r1,chen2025learning,sun2025zerosearch,li2025webthinker}. These agents can autonomously and iteratively invoke external search tools, thereby addressing more challenging knowledge-intensive problems that demand adaptive retrieval and in-depth reasoning. Current research on search agents has made considerable progress in optimizing the reasoning paradigms of search agents through methods like prompt engineering~\cite{li2025search} and fine-tuning \cite{jinsearch,song2025r1,chen2025learning,sun2025zerosearch}. However, they often overlook the quality of intermediate search queries during reasoning, yet low-quality queries can lead to unexpected retrieval results or even derail the entire trajectory. Figure~\ref{fig:overall} illustrates how minor inaccuracies in an intermediate search query (e.g., omitting `actor') can lead a search agent to retrieve and accept unexpected information, ultimately resulting in an incorrect answer. This highlights the critical role that search query quality plays in the deep information-seeking process. Some studies~\cite{wang2025stepsearch,zhang2025process,deng2025atom,xu2025hybrid} have attempted to incorporate process rewards into search agent training. However, they tend to focus more on shaping better reasoning behavior rather than improving the quality of intermediate search queries, and existing efforts~\cite{wang2025stepsearch} on intermediate search queries remain preliminary and ineffective. Furthermore, research~\cite{jiang2025qagent,tao2025webleaper} has shown that existing training paradigms often prioritize information utilization, persistently neglecting the optimization of retrieval patterns. This undoubtedly impedes the search agent's ability to achieve deep and reliable information retrieval, thereby compromising its overall effectiveness. Such issues highlight the need for methods that specifically focus on optimizing query quality during training.

In this work, we present \textbf{\mymodel{}}, a framework that optimizes search query quality through the guidance of process rewards, thereby enhancing the deep information-seeking capabilities of search agents. Specifically, \mymodel{} incorporates two key mechanisms: \textbf{(1) Process rewards}: To provide fine-grained supervision for the quality of each search query, we introduce Dual-Level Credit Assessment, which comprises two complementary components. The first one is a rule-based assessment for query novelty, which detects redundancy by checking whether the retrieved documents contain excessive overlap with previous rounds. The second one is a model-based evaluation for query usefulness, which judges whether the query intent is necessary and whether the retrieved results provide the expected answer. This mechanism outputs both numerical scores and textual feedback, which serve as guidance for subsequent query refinement. \textbf{(2) Query refinement}: To further promote the optimization of query generation during training, the agent first generates a complete search trajectory, then identifies low-quality search rounds according to the numerical scores from the process rewards. Subsequently, we employ a model to refine those queries under the textual guidance provided by the process rewards, after which the search agent continues generating from the refined queries. To improve the efficiency of query assessment and refinement, a smaller model is trained for scoring and refinement, reducing computational cost while maintaining effectiveness.

Building on the foundation of the two mechanisms, we introduce a three-stage curriculum learning framework. The framework guides the search agent through a progression from imitation and alignment to generalization, enabling it to progressively internalize the ability to enhance query quality under the guidance of process rewards. \textbf{(1) Query Quality Screened Imitation Learning}: The initial stage leverages Supervised Fine-Tuning (SFT) to guide the search agent during its early learning of information retrieval and utilization. The training data is filtered based on both final answer correctness and query quality measured by the process rewards. It ensures the model to learn from trajectories that not only lead to correct answers but also maintain high-quality search processes. \textbf{(2) Query Generation Alignment}: In this stage, the search agent cultivates advanced query generation capabilities through Direct Preference Optimization (DPO). We employ the query refinement mechanism to generate comparative data, with process rewards and outcome rewards jointly defining which trajectories are of higher quality. \textbf{(3) Query-Aware Policy Optimization}: The final stage utilizes Reinforcement Learning (RL) to further strengthen its integrated capabilities of information retrieval and utilization. During the rollout phase, the query refinement mechanism is employed, with the process rewards incorporated into the reward function.

To thoroughly assess the capabilities of \mymodel{}, we perform experiments on four challenging knowledge-intensive tasks and two web exploration tasks. Experimental results indicate that \mymodel{} consistently surpasses all baselines in overall performance and exhibits strong generalization to open-web settings. Additionally, we perform a range of ablation studies and quantitative analyses to comprehensively validate \mymodel{}'s effectiveness. Our findings highlight the critical contribution of our two key mechanisms and three curriculum learning stages, as well as their superiority in terms of search efficiency, search query quality, and other dimensions. 

To summarize, the primary contributions of this study include:

(1) We present a pioneering focus that optimizes the quality of intermediate search queries through process reward guidance, thereby improving the information-seeking ability of search agents.

(2) We propose \mymodel{}, a framework that incorporates two key mechanisms: process rewards and query refinement, to enable process reward-guided search refinement.

(3) We design a three-stage, query-oriented curriculum learning framework that guides the agent through a progression from imitation and alignment to generalization, progressively internalizing the ability to improve query quality.

(4) Experiments across six challenging benchmarks demonstrate that \mymodel{} consistently surpasses existing baselines, and further quantitative analyses confirm significant improvements in both search efficiency and query quality.

\section{Related Works}

\subsection{LLM-based Search Agents}
LLMs have demonstrated strong performance across various tasks \cite{achiam2023gpt,touvron2023llama,touvron2023llama2,brown2020language,chowdhery2023palm}, yet challenges like hallucinations \cite{ji2023survey,rawte2023survey} and static parametric knowledge remain. Nowadays, LLM-based search agents have emerged as a promising solution \cite{li2025search,jinsearch,song2025r1,chen2025learning,sun2025zerosearch,li2025webthinker}. This advanced paradigm enables models to autonomously and iteratively invoke external tools, effectively tackling challenging knowledge-intensive problems. Research on search agents has progressed through methods including prompt engineering and fine-tuning. Early prompt-based methods~\cite{li2025search,lu2025octotools,lei2023instructerc,trivedi2023interleaving} focused on carefully designed prompts and structured workflows to steer the agent’s behavior. However, these methods don't fundamentally enhance the model's underlying capabilities, leading many studies to shift towards fine-tuning-based approaches. A prominent line of work has demonstrated that SFT~\cite{fang2025cognitive,jiang2025rag,li2025retrollm,dong2025understand} on expert trajectories enables agents to learn through imitation and yields promising performance. Building upon this foundation, recent studies~\cite{jinsearch,song2025r1,chen2025learning,sun2025zerosearch,dong2025tool} have employed RL to further advance search agent capabilities. However, existing methods tend to overlook intermediate search query quality, which can lead to unexpected retrieval results or even derail the entire trajectory. Moreover, research~\cite{jiang2025qagent} indicates that current training paradigms tend to prioritize information utilization, which can lead to stagnation in information retrieval abilities. Thus, we present a framework designed to optimize the quality of intermediate search queries under the guidance of process rewards, thereby enhancing the overall performance of search agents.

\subsection{Process Rewards in RL}

Recent advancements in RL have achieved significant success in large reasoning models~\cite{chusft,teamkimi,team2024qwq}, and have also demonstrated effectiveness in enhancing the performance of LLM-based search agents~\cite{li2025search,jinsearch,song2025r1,chen2025learning,sun2025zerosearch,li2025webthinker,dong2025tool}. However, reward signals based solely on final outcomes often result in sparse feedback in multi-round search tasks, providing insufficient guidance for intermediate steps and leading to unstable and inefficient policy optimization~\cite{zhang2025landscape}. To overcome this limitation, recent studies~\cite{wang2025stepsearch,zhang2025process,deng2025atom,xu2025hybrid} have explored the use of process-based rewards. Some approaches employ Monte Carlo Tree Search to estimate intermediate actions' value~\cite{zhang2025process,leng2025decex}, while others rely on a corpus of annotated golden steps or intermediate information to compute rewards based on alignment with this reference~\cite{wang2025stepsearch,wang2025beyond,wang2025information,zeng2025reinforcing,zhao2025r}. Still others leverage external reward models to provide fine-grained evaluation for each step~\cite{xu2025hybrid,xiong2025rag,deng2025atom,xu2025beyond,wu2025hiprag}. These approaches have proven effective in enhancing the effectiveness and stability of RL training. Yet, most of these approaches tend to focus primarily on the quality of the reasoning process rather than the quality of intermediate search queries~\cite{deng2025atom,xu2025beyond}, with existing efforts on intermediate search queries remaining preliminary and ineffective~\cite{wang2025stepsearch}. In this context, our process rewards mechanism provides fine-grained supervision for query quality through Dual-Level Credit Assessment, playing a central role in the query-oriented training framework.

\begin{figure*}
  \centering
  \includegraphics[width=\linewidth]{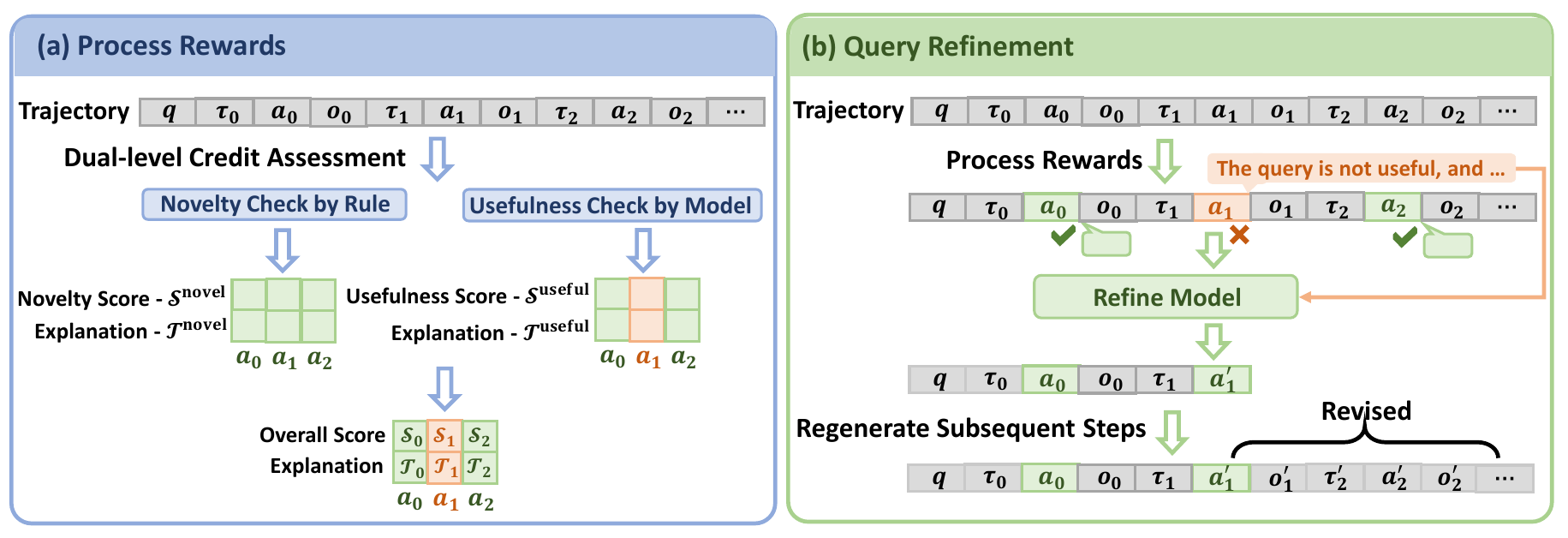}
  \caption{An overview of the two key mechanisms in \mymodel{}: the process rewards (a) and the query refinement (b).}
  \label{fig:mechanism}
\end{figure*}

\section{Preliminaries}

\subsection{Task Formulation}
We adopt ReAct~\cite{yao2022react} as the framework for the search agent. Given a user query \( q \), the search agent, guided by an LLM policy \( \pi_{\theta} \), interacts with an external search tool through several iterations of Thought-Action-Observation to gather information and ultimately generate an answer. Specifically, during each iteration, the search agent starts by engaging in thinking to generate a ``Thought'' according to the existing context. It then produces the next ``Action'', which involves querying the search tool. The agent subsequently waits for the environment to return the ``Observation'', consisting of the Top-K retrieved document fragments for the search query. The iteration concludes when the search agent has gathered sufficient information required to address the user's question and selects the ``final answer'' as the action. A complete trajectory over \( T \) iterations is denoted as: 
\begin{equation}
H_T = (q, \tau_0, a_0, o_0, \dots, \tau_i, a_i, o_i, \dots, \tau_T, a_T).
\end{equation}
Here, \( \tau_i \), \( a_i \), and \( o_i \) correspond to the Thought, Action, and Observation of the \( i \)-th iteration. In iteration \( t \), the LLM policy \( \pi_{\theta}(a, t | H_{t-1}) \) produces the thought \( \tau_t \) and action \( a_t \), which is conditioned on the entire history of prior context \( H_{t-1} \).

\subsection{Agentic Reinforcement Learning}

\paragraph{\textbf{Policy Optimization}} In the context of Agentic RL~\cite{zhang2025landscape}, Group Relative Policy Optimization (GRPO)~\cite{shao2024deepseekmath} is typically employed for policy optimization.  In our approach, we also employ GRPO during the Query Aware Policy Optimization stage, with a specific focus on the augmentation of the rollout and reward modules to optimize the quality of intermediate search queries. Specifically, GRPO optimizes the policy model through maximization of the objective function below: 
\label{sec:policy optimization}
\begin{align}
J_{\text{GRPO}}(\theta) = \; & \mathbb{E}_{(q,a) \sim \mathcal{D}, \{o_i\} \sim \pi_{\theta_{\text{old}}}(\cdot \mid q)} \Biggl[ \frac{1}{G} \sum_{i=1}^G \frac{1}{|o_i|} \sum_{t=1}^{|o_i|} \min\biggl( r_t(\theta) \hat{A}_i,\; \nonumber \\
& \text{clip}\left(r_t(\theta), 1 - \epsilon, 1 + \epsilon\right) \hat{A}_i \biggr) - \beta\, D_{\text{KL}}(\pi_\theta \| \pi_{\text{ref}}) \Biggr].
\end{align}
In this formulation, for each input pair \((q, a)\) drawn from the dataset \(\mathcal{D}\), \(G\) trajectories \(\{o_i\}_{i=1}^G\) are generated from the old policy \(\pi_{\theta_{\text{old}}}(\cdot \mid q)\). The importance weight \(r_t(\theta)\) is defined as:
\begin{equation}
r_t(\theta) = \frac{\pi_\theta(o_{i,t} \mid q, o_{i,<t})}{\pi_{\theta_{\text{old}}}(o_{i,t} \mid q, o_{i,<t})}.
\end{equation}
The normalized advantage score \(\hat{A}_i\)  is denoted as:
\begin{equation}
\hat{A}_i = \frac{r_i - \text{mean}(\{r_j\}_{j=1}^G)}{\text{std}(\{r_j\}_{j=1}^G)}.
\end{equation}
Here, \(r_i\) denotes the scalar reward for the \(i\)-th rollout. Furthermore, agentic RL typically masks observations originating from the external environment during loss computation, thereby preventing unstable training.

\paragraph{\textbf{Reward Design}} As discussed earlier, in agentic RL, each rollout corresponds to a scalar reward \( r \). Prior research \cite{jinsearch,song2025r1,chen2025learning,sun2025zerosearch} predominantly relies on combining two key types of rewards: the outcome reward \( r_{\text{outcome}} \), reflecting the trajectory's answer correctness, and the format reward \( r_{\text{format}} \), assessing the trajectory's structural correctness. These rewards are typically weighted and combined using a simple hyperparameter \( \lambda \) as follows:
\begin{equation}
r = r_{\text{outcome}} + \lambda \cdot r_{\text{format}}.
\end{equation}
In some recent works \cite{deng2025atom,xu2025hybrid,xu2025beyond}, process rewards have been incorporated into the reward function to provide fine-grained feedback on intermediate steps. The reward function is then extended to include the process rewards, with a composite reward incorporating both the outcome reward and the process rewards, while the format reward is weighted by a hyperparameter: 
\begin{equation}
r = r_{\text{composite}} + \lambda \cdot r_{\text{format}}.\label{eqn:reward_r}
\end{equation}
Here, \( r_{\text{composite}} \) is computed as the aggregation of multiple step-wise process rewards and the final outcome reward $r_{\text{outcome}}$:
\begin{equation}
r_{\text{composite}} = f(r^{\text{process}}_{1}, r^{\text{process}}_{2}, \dots, r^{\text{process}}_{n}, r_{\text{outcome}}),\label{eqn:r_composite}
\end{equation}
where \( n \) represents the total steps in the trajectory, and \( r^{\text{process}}_{i} \) denotes the process reward for the \( i \)-th step. The aggregation function \( f \) combines these individual rewards, and its specific form may vary across different works.

\section{Our Method}

\subsection{Overview}

\begin{figure*}
  \centering
  \includegraphics[width=\linewidth]{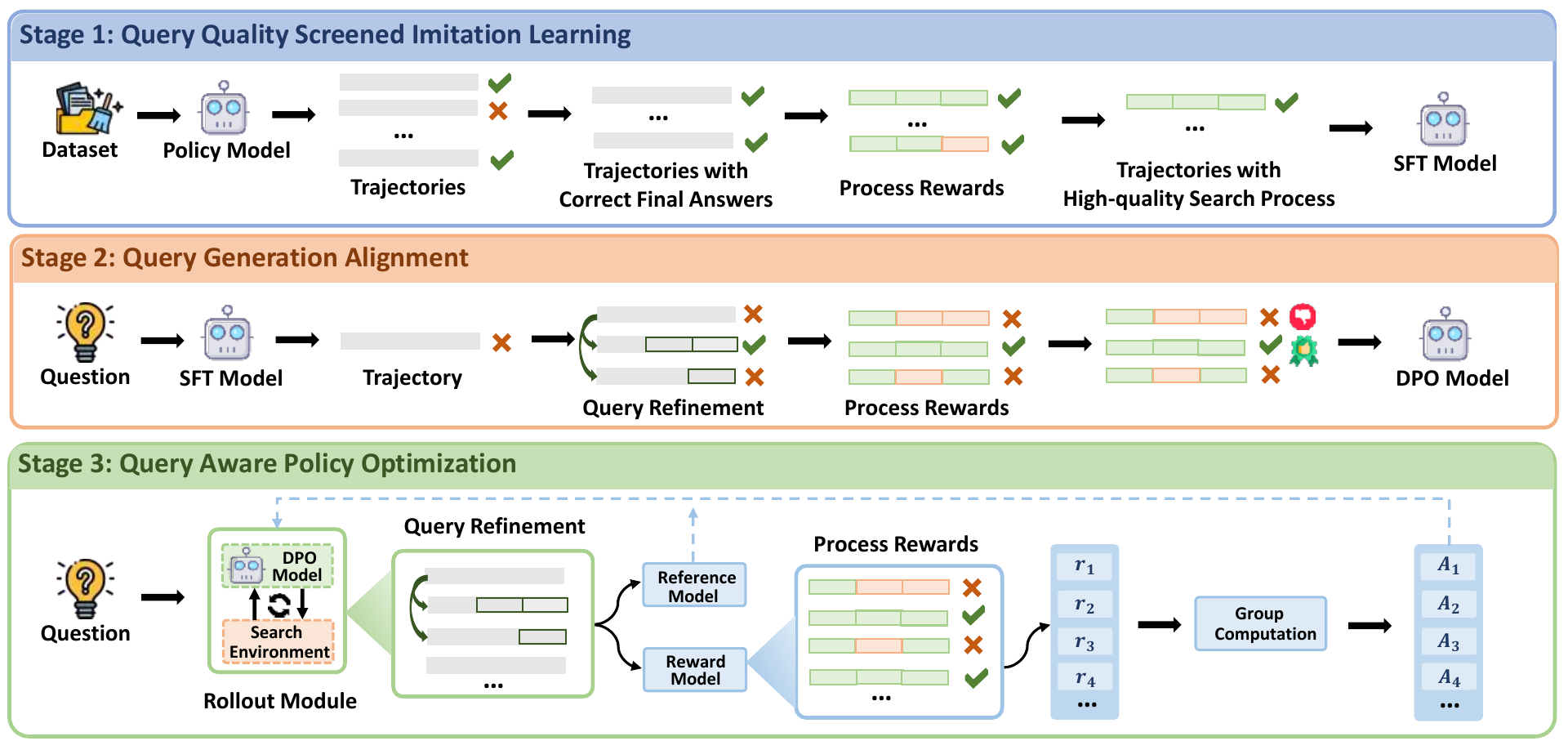}
  \caption{The overall framework of query-oriented three-stage curriculum learning, including Query Quality Screened Imitation Learning, Query Generation Alignment, and Query Generation Alignment.}
  \label{fig:training}
\end{figure*}

We propose \mymodel{}, a framework that enhances search agent performance by optimizing the quality of intermediate search queries through process reward guidance. As illustrated in Figure~\ref{fig:mechanism}, \mymodel{} incorporates two key mechanisms: (1) \textbf{Process rewards} (§\ref{sec:Process Reward for Query Quality}), which provide fine-grained supervision for the quality of each query through Dual-Level Credit Assessment. (2) \textbf{Query refinement} (§\ref{sec:Query Refinement}), which promotes the optimization of query generation by selectively refining low-quality queries and regenerating subsequent search rounds based on these refinements. To further internalize the ability to improve query quality, we propose a three-stage curriculum learning framework (§\ref{sec:Three-stage Curriculum Learning}) built upon these mechanisms. As shown in Figure~\ref{fig:training}, it comprises \textbf{Query Quality Screened Imitation Learning}, \textbf{Query Generation Alignment}, and \textbf{Query Aware Policy Optimization}. Below, we will first introduce the two key mechanisms, followed by a detailed description of the three-stage curriculum learning framework.

\subsection{Process Reward for Assessing Query Quality}
\label{sec:Process Reward for Query Quality}

In this section, we introduce the process rewards mechanism to assess the quality of each query, providing both numerical scores and textual feedback. These outputs guide the subsequent query refinement, and play a key role within the three-stage curriculum learning framework by selecting trajectories with high-quality search processes and providing finer-grained supervision signals, which will be introduced later.

\paragraph{\textbf{Design Principles}} Our assessment of search query quality is guided by a comprehensive set of three fundamental principles: 
\begin{itemize}[leftmargin=1em]
    \item \textbf{Query Novelty}: The query should avoid redundancy with previous queries and introduce novel information.
    \item \textbf{Intent Necessity}: The query's search intent must be necessary for progressing \emph{toward the final answer}.
    \item \textbf{Retrieval Relevance}: The retrieved documents should align with the search intent, effectively containing the expected information or answer.
\end{itemize}
These principles are well-motivated and collectively capture the essential aspects of a high-quality query, while also being readily applicable via either rule-based checks or simple model judgments.

\paragraph{\textbf{Dual-Level Credit Assessment.}} We operationalize these principles through Dual-Level Credit Assessment, which consists of two complementary components. 

\textbf{(1) Rule-based Evaluation:} The first is a rule-based evaluation for query novelty, which identifies redundant queries by measuring the document overlap between the current and previous search rounds. Formally, for the $t$-th step, the novelty score $\mathcal{S}^{\text{novel}}_{t}$ and its corresponding textual explanation $\mathcal{T}^{\text{novel}}_{t}$ are defined as:
\begin{equation}
\left(\mathcal{S}^{\text{novel}}_{t}, \mathcal{T}^{\text{novel}}_{t}\right) =
\begin{cases}
(0, \text{the query is redundant}), & \text{if } O_{t} > K, \\[3pt]
(1, \text{the query is novel}), & \text{if } O_{t} \leq K.
\end{cases}
\end{equation}
Here, $K$ is a threshold hyperparameter, and $O^{t}$ represents the number of documents retrieved at step $t$ that share the same content with those retrieved in any previous step, defined as:
\begin{equation}
O^{t} = \sum_{i=1}^{n} \mathbb{I}(D_{i}^{t} \in \bigcup_{s=0}^{t-1} \bigcup_{j=1}^{n} D_{j}^{s}),
\end{equation}
where $D_{i}^{t}$ refers to the $i$-th document retrieved at step $t$, and $\mathbb{I}(\cdot)$ is the indicator function.

\textbf{(2) Model-based Evaluation:} The second component is model-based evaluation for query usefulness, which assesses the necessity of the query intent and checks whether the retrieved results provide the expected answer. For the $t$-th step, the evaluation score $\mathcal{S}^{\text{useful}}_{t}$ and its corresponding textual explanation $\mathcal{T}^{\text{useful}}_{t}$ are defined as:
\begin{equation}
\mathcal{S}^{\text{useful}}_{t}, \, \mathcal{T}^{\text{useful}}_{t} = \text{LLM}_\text{eval}(q, a, H_t),
\end{equation}
where $\text{LLM}_\text{eval}$ is the model used for evaluation, $q$ is the user's query, $a$ denotes the golden answer, and $H_t$ indicates the trajectory up to step $t$. The score $\mathcal{S}^{\text{useful}}_{t}$ is set to 1 if the query meets the criteria, and 0 otherwise, while the explanation $\mathcal{T}^{\text{useful}}_{t}$ is directly parsed from the model's output. To enhance efficiency, we employ a smaller model fine-tuned via SFT for both scoring and the subsequent query refinement task. Specifically, we input task-specific prompts into a more powerful teacher model and use its outputs as annotation labels. The smaller model is then trained on these prompt-output pairs, enabling it to achieve effective performance at a reduced computational cost. More details about the model will be introduced in Section \ref{sec:implementation details}.

Finally, the overall assessment score $\mathcal{S}_t$ and its corresponding textual explanation $\mathcal{T}_t$ are derived by aggregating the evaluations for query novelty and usefulness. The overall score is determined by a logical conjunction of the component scores: 
\begin{equation}
\mathcal{S}_{t} = 
\begin{cases}
1, & \text{if } \mathcal{S}^{\text{novel}}_{t} = 1 \land \mathcal{S}^{\text{useful}}_{t} = 1, \\
0, & \text{otherwise}.
\end{cases}\label{eq:query_quality_score}
\end{equation}
The final explanation is synthesized by concatenating the textual feedback from both components:
\begin{equation}
\mathcal{T}_{t} = \mathcal{T}^{\text{novel}}_{t} \, \| \, \mathcal{T}^{\text{useful}}_{t},\label{eq:query_quality_desc}
\end{equation}
where $\|$ denotes the concatenation operator.

\subsection{Process Reward-Guided Query Refinement}
\label{sec:Query Refinement}

This section introduces the query refinement mechanism, which is designed to promote the optimization of query generation. It is achieved by systematically identifying and refining low-quality queries, then regenerating subsequent search steps from these refined points. This mechanism serves a pivotal function within the three-stage curriculum learning framework by generating comparative data for training and acting as a rollout strategy.

Formally, this process can be represented as follows. The search agent starts by generating a complete trajectory $H_T$, represented as $(q, \tau_0, a_0, o_0, \dots, \tau_i, a_i, o_i, \dots, \tau_T, a_T)$. Each search query in this trajectory is then evaluated by the process rewards mechanism, yielding a sequence of scores $(\mathcal{S}_0, \mathcal{S}_1, \ldots, \mathcal{S}_{T-1})$ and corresponding textual explanations $(\mathcal{T}_0, \mathcal{T}_1, \ldots, \mathcal{T}_{T-1})$ with Equation~(\ref{eq:query_quality_score}) and (\ref{eq:query_quality_desc}). For each low-quality query $a_i$ where the score $\mathcal{S}_i = 0$, a refinement step is triggered. The refined query $a'_i$ is generated by a language model as follows:
\begin{equation}
a'_i = \text{LLM}_\text{refine}(q, H_i, \mathcal{T}_i).
\end{equation}
Here, $\text{LLM}_\text{refine}$ is the same lightweight SFT-tuned model introduced earlier, $q$ is the user's original query, $H_i$ is the trajectory history up to step $i$, and $\mathcal{T}_i$ is the textual feedback diagnosing the quality issue for the low-quality query $a_i$. The search agent subsequently regenerates the search process from this refined query $a'_i$, yielding a new trajectory $H'_T$, represented as $(q, \tau_0, a_0, o_0, \ldots, \tau_i, a'_i, o'_i, \ldots, \tau'_T, \\a'_T)$. The primary distinction between the initial and revised trajectories originates from the refined query $a'_i$, resulting in a different reward for $a_i$ and $a'_i$, thereby promoting the optimization of query generation within the curriculum learning framework.

Specifically, to enable the model to effectively refine the low-quality search query based on the textual feedback provided by the process rewards mechanism, we distill key empirical insights into a set of actionable guidelines based on a thorough analysis of representative cases: 
\begin{itemize}[leftmargin=1em]
    \item If the textual feedback indicates that the query is redundant or unnecessary, the refined query should serve for a more necessary intent and eliminate redundancy.
    \item If the textual feedback indicates that the retrieved results do not contain the expected information, the model should strategically reformulate the query to better capture the target content. This reformulation may involve switching between a complete semantic question and a keyphrase-based query, or adaptively adding or removing information from the original query.
\end{itemize}

\subsection{Query-Oriented Training Framework}
\label{sec:Three-stage Curriculum Learning}
This section presents a three-stage curriculum learning framework that integrates the two preceding mechanisms, enabling the agent to progressively internalize the ability to improve query quality through a progression from imitation, to alignment, and ultimately to generalization. The following paragraphs detail the three progressive stages: Query Quality Screened Imitation Learning, Query Generation Alignment, and Query Aware Policy Optimization.

\paragraph{\textbf{Stage-1: Query Quality Screened Imitation Learning}} \label{sec:Query Quality Screened Imitation Learning} In this stage, we employ SFT to guide the model in its initial learning of information retrieval and utilization. A critical step in SFT is the selection of high-quality trajectories for training. Following common practice, we begin by selecting trajectories that yield correct final answers and adhere to the proper format, thus guiding the model towards correct patterns from the outset. However, many trajectories, despite yielding correct final answers, contain low-quality intermediate search queries. Learning from such trajectories could lead the model to pick up suboptimal behaviors, thereby impairing its overall performance. To address this, we further leverage process rewards to selectively retain only those trajectories comprised entirely of high-quality intermediate search queries, i.e., $\forall t \in [0,\dots,T], \mathcal{S}_{t}=1$. This ensures that the trajectories comprising our final training dataset $\mathcal{D}$ not only yield correct final answers but also exhibit high-quality intermediate search queries. We then apply the standard SFT objective, which is formulated as:
\begin{equation}
\mathcal{L}_{\text{SFT}}(\theta) = -\mathbb{E}_{(q, y) \sim \mathcal{D}} \left[ \log P_\theta(y \mid q) \right],
\end{equation}
where \(q\) is the user’s
original query, \(y\) is the agent's high-quality response, and \(\theta\) denotes the model parameters.

\paragraph{\textbf{Stage-2: Query Generation Alignment}} \label{sec:Query Generation Alignment} In this stage, the search agent cultivates advanced query generation capabilities through DPO training. Unlike common approaches that directly generate trajectories from scratch, we employ the query refinement mechanism when constructing comparative data. For each user’s query \(q\), the search agent first generates an initial trajectory \(y_0\). Following this, each low-quality query within \(y_0\) is refined and the search agent regenerates subsequent search steps from the refined query, producing a sequence of trajectories  \(y_1, \dots, y_n\), where \(n\) is the number of low-quality queries. This process ensures that for a given input \(q\), the key differences among the candidate trajectories \(y_0, y_1, \dots, y_n\) originate specifically from the refined queries, thereby directly promoting the optimization of query generation.

Next, for each user’s query \(q\), we choose one positive sample \(y_w\) and one negative sample \(y_l\) among the corresponding candidate trajectories \(y_0, y_1, \dots, y_n\). Diverging from approaches that rely only on the correctness of the final answer, our selection criteria incorporate \emph{both the final-answer correctness and the quality of intermediate search queries}, guided by the following principles: 
\begin{itemize}[leftmargin=1em]
    \item A trajectory with a correct final answer is preferred over one with an incorrect answer.
    \item Among trajectories with correct final answers, those with fewer low-quality (i.e., $\mathcal{S}_t=0$) queries are preferred.
    \item Among trajectories with incorrect final answers, those containing more high-quality (i.e., $\mathcal{S}_t=1$) queries are preferred.
\end{itemize}
We then optimize the model using the standard DPO objective:
\begin{align}
\mathcal{L}_{\text{DPO}}(\theta) = & -\mathbb{E}_{(q, y_w, y_l) \sim \mathcal{D}} \bigg[ \log \sigma \left( \beta \log \frac{\pi_\theta(y_w \mid q)}{\pi_{\text{ref}}(y_w \mid q)} \right. \nonumber \\
& \left. \hspace{2.5cm} - \beta \log \frac{\pi_\theta(y_l \mid q)}{\pi_{\text{ref}}(y_l \mid q)} \vphantom{\frac{\pi_\theta(y_w \mid q)}{\pi_{\text{ref}}(y_w \mid q)}} \right) \bigg].
\end{align}
Here, \(q\) is the user’s
original query, \(y_w\) is the positive sample,  \(y_l\) is the negative sample, \(\beta\) represents the hyperparameter, \(\sigma\) refers to the sigmoid function, \(\theta\) is the model parameters, and \(\pi_{\text{ref}}\) indicates the reference model, which is initialized to \(\pi_\theta\) and kept frozen during training.

\paragraph{\textbf{Stage-3: Query Aware Policy Optimization}} \label{sec:Query Aware Policy Optimization} In the final stage, we further enhance the search agent's integrated capabilities of information retrieval and utilization through Query Aware Policy Optimization. Specifically, we train it on a curated set of challenging questions that remained unresolved after multiple sampling trials. Unlike the standard GRPO algorithm that generates $G$ independent trajectories from scratch, our method employs the query refinement mechanism as its rollout strategy. For each user's query, the search agent first generates an initial trajectory $y_0$ and then expands it into ${y_0, y_1, \dots, y_n}$ through sequential refinement and regeneration. Different from the Query Generation Alignment stage, we retain at most $M$ trajectories from this set to avoid too many trajectories sharing a common prefix, thereby ensuring behavioral diversity and promoting the holistic improvement of the agent's capabilities. If the total number of trajectories collected remains less than $G$, we repeat this generation-and-expansion process, until a complete set of $G$ trajectories is obtained.

For reward design, we integrate process supervision into the reward function. Following Eq. (\ref{eqn:reward_r}), our reward function is:
\begin{equation}
r = r_{\text{composite}} + \lambda \cdot r_{\text{format}},
\end{equation}
where \(\lambda\) is a weighting coefficient, \(r_{\text{format}} \in \{0, 1\}\) indicates the correctness of the output format, and \(r_{\text{composite}}\) defined in Eq. (\ref{eqn:r_composite}) integrates both outcome and process reward as follows:
\begin{equation}
r_{\text{composite}} =
\begin{cases}
\max(r_{\text{outcome}} - \gamma \cdot n_{\text{wrong}}, \, \phi_{\text{min}}), & r_{\text{outcome}} = 1, \\
\min(r_{\text{outcome}} + \gamma \cdot n_{\text{correct}}, \, \phi_{\text{max}}), & r_{\text{outcome}} = 0.
\end{cases}
\end{equation}
Here, \(r_{\text{outcome}} \in \{0, 1\}\) denotes the final answer's correctness, \(n_{\text{wrong}}\) and \(n_{\text{correct}}\) represent the number of low- (i.e., $\mathcal{S}_t=0$) and high-quality (i.e., $\mathcal{S}_t=1$) queries respectively, \(\gamma\) is a scaling factor for process rewards, and \(\phi_{\text{min}}, \phi_{\text{max}}\) bound the influence of process rewards. This reward design incentivizes the agent not only to prioritize final answer correctness but also to refine its search process by reducing low-quality queries in successful trajectories. Moreover, even when unable to provide a final correct answer, the agent is motivated to generate more high-quality queries that may progressively approach the solution. We then optimize the model using the standard GRPO objective introduced in Section~\ref{sec:policy optimization}.

\section{Experiments}

\begin{table*}[!t]
\centering
\caption{Performance comparison of \mymodel{} and existing approaches on four knowledge-intensive benchmarks, with \textbf{bold} for the best and \underline{underlined} for the runner-up. Numbers in () indicate the improvement compared with the runner-up.}
\label{tab:overall result}
\renewcommand{\arraystretch}{1}
\begin{tabular}{
    p{2.3cm}
    *{2}{p{1.2cm}}
    *{2}{p{1.0cm}}
    *{6}{p{1.2cm}}
}
\toprule
\multirow{2}{*}{\textbf{Method}}
  & \multicolumn{2}{c}{\textbf{2WikiMQA}}
  & \multicolumn{2}{c}{\textbf{HotpotQA}}
  & \multicolumn{2}{c}{\textbf{Bamboogle}}
  & \multicolumn{2}{c}{\textbf{Musique}}
  & \multicolumn{2}{c}{\textbf{Average}} \\
\cmidrule(lr){2-3} \cmidrule(lr){4-5} \cmidrule(lr){6-7} \cmidrule(lr){8-9} \cmidrule(lr){10-11}
  & EM & F1
  & EM & F1
  & EM & F1
  & EM & F1
  & EM & F1 \\
\midrule

\multicolumn{11}{l}{\textit{\textbf{Prompt-based Approaches}}} \\
Direct Inference & 19.3 & 24.7 & 14.6 & 24.5 & 4.0 & 11.6 & 2.3 & 7.9 & 10.1 & 17.2\\
CoT & 18.1 & 24.9 & 12.8 & 24.0 & 14.4 & 25.5 & 2.2 & 7.8 & 11.9 & 20.6\\
IRCoT & 20.0 & 27.2 & 19.3 & 28.0 & 16.8 & 25.9 & 5.8 & 12.7 & 15.5 & 23.5\\
RAG & 22.5 & 31.4 & 24.3 & 36.7 & 7.2 & 17.2 & 4.5 & 12.2 & 14.6 & 24.4\\
Search-o1 & 20.9 & 29.4 & 22.0 & 33.6 & 28.8 & 36.1 & 5.1 & 12.6 & 19.2 & 27.9\\
\midrule
\multicolumn{11}{l}{\textit{\textbf{RL Approaches with Outcome Rewards}}} \\
ReSearch & 29.4 & 36.7 & 28.5 & 40.8 & 12.8 & 22.9 & 10.0 & 17.3 & 20.2 & 29.4\\
ZeroSearch & 29.2 & 36.5 & 27.5 & 39.1 & 14.4 & 25.4 & 10.4 & 18.2 & 20.4 & 29.8\\
R1-Searcher & 29.8 & 37.1 & 27.0 & 38.7 & 31.2 & 39.2 & 8.0 & 16.4 & 24.0 & 32.9\\
Search-R1 & 27.3 & 35.5 & 31.9 & 41.1 & 29.4 & 38.8 & 9.3 & 16.6 & 24.5 & 33.0\\
\midrule
\multicolumn{11}{l}{\textit{\textbf{RL Approaches with Process Rewards}}} \\
ReasonRag & \underline{36.5} & \underline{43.2} & 32.2 & 41.7 & 30.4 & 39.1 & 11.3 & 18.6 & 27.6 & 35.7\\
PPR & 33.7 & 41.8 & \underline{38.1} & \underline{50.3} & 31.2 & 39.4 & 14.7 & 22.0 & 29.4 & 38.4\\
StepSearch & 32.1 & 38.9 & 35.1 & 45.9 & \underline{36.8} & \underline{48.4} & \underline{16.6} & \underline{24.9} & \underline{30.1} & \underline{39.5}\\
\mymodel{}~(Ours) & \textbf{45.3}(\textcolor{blue}{$\uparrow$24\%}) & \textbf{52.3}(\textcolor{blue}{$\uparrow$21\%}) & \textbf{40.7}(\textcolor{blue}{$\uparrow$7\%}) & \textbf{52.4}(\textcolor{blue}{$\uparrow$4\%}) & \textbf{44.8}(\textcolor{blue}{$\uparrow$22\%}) & \textbf{56.1}(\textcolor{blue}{$\uparrow$16\%}) & \textbf{19.1}(\textcolor{blue}{$\uparrow$15\%}) & \textbf{27.8}(\textcolor{blue}{$\uparrow$12\%}) & \textbf{37.5}(\textcolor{blue}{$\uparrow$25\%}) & \textbf{47.2}(\textcolor{blue}{$\uparrow$19\%})\\
\bottomrule
\end{tabular}
\end{table*}

\subsection{Experimental Setup}

\paragraph{\textbf{Dataset}} We comprehensively assess \mymodel{}'s performance through experiments on two types of benchmarks: (1) knowledge-intensive tasks, including 2WikiMultihopQA \cite{ho2020constructing}, HotpotQA \cite{yang2018hotpotqa}, Bamboogle \cite{press2023measuring}, and Musique \cite{trivedi2022musique}, and (2) web exploration tasks, including GAIA \cite{mialon2023gaia} and WebWalker \cite{wuwebwalker}.

\paragraph{\textbf{Metrics}} For a consistent comparison with previous studies, we use the widely adopted Exact March (EM) and word-level F1 score to assess the answers' correctness. To assess search efficiency, we follow prior work \cite{chen2025toward} and employ the Search Efficiency metric, defined as: $ S_E = \frac{1}{N} \sum_{i=1}^{N} \frac{F_i}{T_i} $. Here, \(N\) represents the dataset size, \(F_i\) denotes the F1 score for sample \(i\), and \(T_i\) represents the search call count for sample \(i\). Additionally, to assess search query quality, we introduce the Search Quality metric, defined as: $ S_Q = \frac{1}{N} \left( C_{\text{perfect}} + C_{\text{partial}} \right) $ where \(N\) represents the dataset size, \(C_{\text{perfect}}\) denotes the number of samples where the final answer is correct and all intermediate search queries are of high quality, and \(C_{\text{partial}}\) denotes the number of samples where the final answer is incorrect but the trajectory contains high-quality intermediate search queries. In particular, we define the Perfect Rate as $ \frac{1}{N}  C_{\text{perfect}} $ and the Partial Rate as $ \frac{1}{N}  C_{\text{partial}} $, which contribute to the overall Search Quality metric from two different aspects.

\paragraph{\textbf{Baselines}} We compare \mymodel{} with several representative baselines, which are classified into three categories: (1) prompt-based approaches, including Direct Inference, CoT \cite{wei2022chain}, IRCoT \cite{trivedi2023interleaving}, RAG \cite{lewis2020retrieval}, and Search-o1 \cite{li2025search}. (2) RL approaches with outcome rewards, including ReSearch \cite{chen2025learning}, ZeroSearch \cite{sun2025zerosearch}, R1-Searcher \cite{song2025r1}, and Search-R1 \cite{jinsearch}. (3) RL approaches with process rewards, including PPR \cite{xu2025hybrid}, ReasonRag \cite{zhang2025process}, and StepSearch \cite{wang2025stepsearch}.

\paragraph{\textbf{Implementation Details}} 
\label{sec:implementation details}
Qwen2.5-3B-Instruct serves as the base model in \mymodel{} and other baselines. For local search, we utilize the 2018 Wikipedia dump \cite{karpukhin2020dense} provided by FlashRAG \cite{jin2025flashrag} as the knowledge corpus, and employ E5-base-v2 \cite{wang2022text} as the retriever to obtain top 5 documents. For web search, the Serper API is employed to retrieve the top 10 document snippets. Our training is conducted under the LLaMA-Factory and VERL frameworks, using Asearcher-Base as the training dateset. Additionally, to improve efficiency, we train a smaller student model, Qwen2.5-3B-Instruct, to perform scoring and query refinement, with training labels annotated by the teacher model, Qwen3-32B.

\subsection{Main Result}

\begin{table}
\centering
\caption{Performance comparison of \mymodel{} and baselines on web exploration tasks, with \textbf{bold} for the best.}
\label{tab:web_exp_results}
\renewcommand{\arraystretch}{1}
\begin{tabular}{
    p{1.7cm}
    *{2}{>{\centering\arraybackslash}p{0.7cm}}
    *{2}{>{\centering\arraybackslash}p{0.7cm}}
    *{2}{>{\centering\arraybackslash}p{0.7cm}}
}
\toprule
\multirow{2}{*}{\textbf{Method}}
  & \multicolumn{2}{c}{\textbf{GAIA}}
  & \multicolumn{2}{c}{\textbf{WebWalker}}
  & \multicolumn{2}{c}{\textbf{Average}} \\
\cmidrule(lr){2-3} \cmidrule(lr){4-5} \cmidrule(lr){6-7}
  & EM & F1
  & EM & F1
  & EM & F1 \\
\midrule

Search-o1 & 4.7 & 8.0 & 6.3 & 18.3 & 5.5 & 13.2\\
Search-R1 & 6.3 & 9.8 & 7.5 & 21.6 & 6.9 & 15.7\\
StepSearch & 9.4 & 12.5 & 9.1 & 25.8 & 9.3 & 19.2\\
\mymodel{} & \textbf{13.4} & \textbf{16.7} & \textbf{11.5} & \textbf{31.0} & \textbf{12.5} & \textbf{23.9}\\
\bottomrule
\end{tabular}
\end{table}

\begin{table*}[!t]
\centering
\caption{Ablation study results for the two core mechanisms in \mymodel{} across all three stages of curriculum learning training framework, with \textbf{bold} for the best results of each stage.}
\label{tab:ablation study}
\renewcommand{\arraystretch}{1}
\begin{tabular}{
    p{3.3cm}
    *{2}{>{\centering\arraybackslash}p{1cm}}
    *{2}{>{\centering\arraybackslash}p{1cm}}
    *{2}{>{\centering\arraybackslash}p{1cm}}
    *{2}{>{\centering\arraybackslash}p{1cm}}
    *{2}{>{\centering\arraybackslash}p{1cm}}
}
\toprule
\multirow{2}{*}{\textbf{Method}}
  & \multicolumn{2}{c}{\textbf{2WikiMQA}}
  & \multicolumn{2}{c}{\textbf{HotpotQA}}
  & \multicolumn{2}{c}{\textbf{Bamboogle}}
  & \multicolumn{2}{c}{\textbf{Musique}}
  & \multicolumn{2}{c}{\textbf{Average}} \\
\cmidrule(lr){2-3} \cmidrule(lr){4-5} \cmidrule(lr){6-7} \cmidrule(lr){8-9} \cmidrule(lr){10-11}
  & EM & F1
  & EM & F1
  & EM & F1
  & EM & F1
  & EM & F1 \\
\midrule

\multicolumn{11}{l}{\textit{\textbf{Stage 1}}} \\
\mymodel{}  & \textbf{38.2} & \textbf{45.3} & \textbf{35.3} & \textbf{45.5} & \textbf{38.4} & \textbf{51.0} & \textbf{14.7} & \textbf{21.6} & \textbf{31.7} & \textbf{40.9}\\
\quad w/o process rewards & 33.8 & 40.6 & 32.5 & 42.8 & 36.0 & 48.7 & 12.6 & 20.8 & 28.7& 38.2\\
\midrule
\multicolumn{11}{l}{\textit{\textbf{Stage 2}}} \\
\mymodel{} & \textbf{41.4} & \textbf{48.7} & \textbf{37.9} & \textbf{48.5} & \textbf{39.2} & \textbf{51.8} & \textbf{15.4} & \textbf{23.6} & \textbf{33.5} & \textbf{43.2} \\
\quad w/o process rewards & 40.2 & 47.4 & 36.5 & 47.2 & 37.6 & 50.1 & 14.4 & 22.3 & 32.2 & 41.8\\
\quad w/o query refinement & 39.2 & 46.7 & 35.6 & 46.1 & 36.0 & 49.6 & 14.6 & 22.9 & 31.4 & 41.3 \\
\midrule
\multicolumn{11}{l}{\textit{\textbf{Stage 3}}} \\
\mymodel{} & \textbf{45.3} & \textbf{52.3} & \textbf{40.7} & \textbf{52.4} & \textbf{44.8} & \textbf{56.1} & \textbf{19.1} & \textbf{27.8} & \textbf{37.5} & \textbf{47.2}\\
\quad w/o process rewards  & 43.3 & 50.6 & 40.0 & 51.5 & 39.2 & 51.6 & 17.9 & 26.7 & 35.1 & 45.1\\
\quad w/o query refinement & 44.1 & 51.2 & 39.9 & 51.6 & 41.6 & 54.2 & 17.5 & 26.4 & 35.8 & 45.9\\
Standard GRPO  & 43.6 & 50.8 & 39.0 & 50.6 & 40.8 & 52.5 & 15.9 & 24.8 & 34.8 & 44.7\\
\bottomrule
\end{tabular}
\end{table*}

\paragraph{\textbf{Overall Performance}} Table~\ref{tab:overall result} presents the main results, demonstrating that \mymodel{} consistently surpasses existing approaches across four datasets, and yielding several important insights.

\textbf{(1) Prompt-based approaches exhibit limited performance.} Direct Inference and CoT, which are based entirely on the model's internal knowledge, achieve an average EM of only around 10\%, highlighting the inherent challenges of LLMs, including hallucinations and static parametric knowledge. RAG and IRCoT yield a gain of around 5\% in average EM, demonstrating the necessity of integrating external knowledge. Among prompt-based methods, Search-o1 attains the highest performance, reaching an average EM score of 19.2\%, reflecting the effectiveness of search agents. However, it still lags behind other fine-tuning-based approaches.

\textbf{(2) Incorporating process rewards effectively enhances RL training.} While outcome-driven RL methods such as ReSearch, Search-R1, and R1-Searcher improve performance over prompt-based approaches, indicating the benefits of RL for LLM-based search agents, they often remain inferior to RL approaches that integrate both outcome and process rewards by a margin of around 5\% in both average EM and F1 score. This underscores that reward signals based solely on final outcomes result in sparse feedback. Such sparse feedback provides insufficient guidance for intermediate steps and leads to unstable optimization, thereby highlighting the critical importance of fine-grained supervision.

\textbf{(3) Optimizing the quality of intermediate search queries significantly improves overall performance.} Existing methods often overlook the quality of intermediate search queries, which can lead to stagnation in information retrieval abilities. By explicitly optimizing the quality of intermediate search queries under the guidance of process rewards, \mymodel{} enhances the search agent's overall effectiveness, achieving more than 7\% improvement in both average EM and F1 score compared with other process-supervised RL methods.

\paragraph{\textbf{Generalization to Web Search Scenarios}} As discussed earlier, \mymodel{} is trained solely on Wikipedia-based local search. To evaluate its generalization ability to web search, we test it against several baseline models on two demanding web exploration tasks, GAIA and WebWalker. As demonstrated in Table~\ref{tab:web_exp_results}, \mymodel{} surpasses existing approaches across both datasets, achieving an average F1 score increase of nearly 5\%. This indicates that while \mymodel{} optimizes the quality of intermediate search queries in the local search setting, it also exhibits strong generalization capabilities in the web search environment, maintaining robust performance despite the difference in settings.
\begin{figure}[t]
  \centering
  \includegraphics[width=\linewidth]{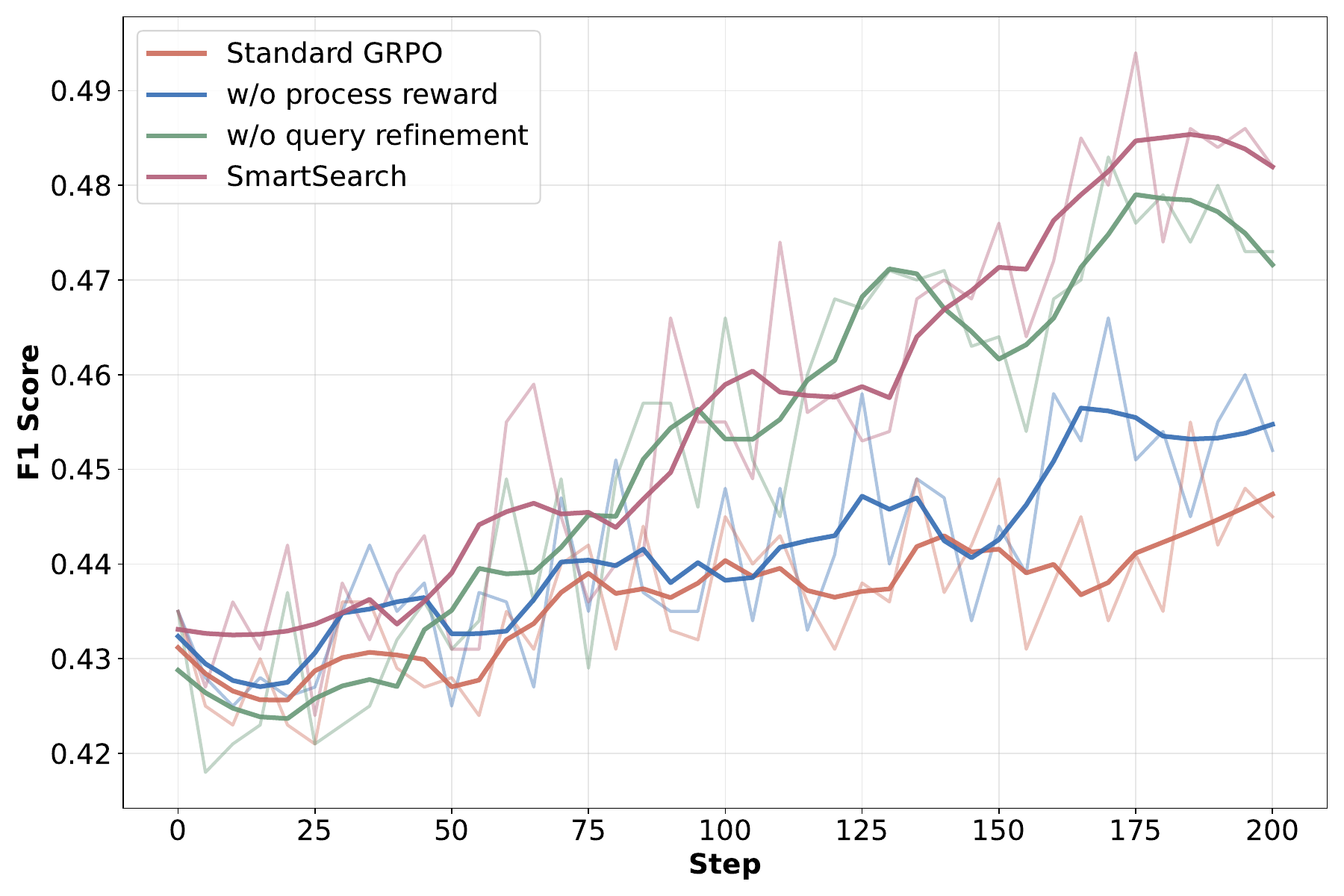}
  \caption{F1 score training dynamics for different algorithms.}
  \label{fig:ablation study}
\end{figure}
\subsection{Ablation Study}

To further examine the impact of \mymodel{}'s two key mechanisms—process rewards and query refinement, we conduct extensive ablation studies across all three training stages. The results are summarized in Table \ref{tab:ablation study}.

For Stage 1, we compare our configuration with a baseline that filters the training data exclusively according to whether the final answer is correct. The results indicate that incorporating query-quality filtering enables the model to achieve superior performance with only 60\% of the training data, highlighting the importance of learning from trajectories with high-quality search processes.

For Stage 2, we compare our method with two alternatives: (1) directly generating full trajectories without query refinement, and (2) determining preference based exclusively on the final answer correctness. Ablation results underscore the essential contribution of both mechanisms in this satge, particularly the query refinement mechanism, underscoring the significance of promoting the optimization of query generation.

For Stage 3, we compare our algotithm with three variants: a GRPO baseline, a version that only incorporates the process rewards into the reward function, and a version that only applies the query refinement mechanism during rollout. As depicted in Figure \ref{fig:ablation study}, we demonstrate F1 score curves of various RL algorithms during training. The results demonstrate that our algorithm reliably outperforms all alternatives. Notably, integrating process rewards into the reward function yields significant gains, illustrating the crucial role of fine-grained supervision for the quality of each query.

\subsection{Quantitative Analyses}
To comprehensively assess the effectiveness of the \mymodel{} framework, we perform multiple quantitative experiments. The following analyses demonstrate its superiority in four key aspects: intermediate query quality, search efficiency, the effectiveness of the process reward model, and the effectiveness-efficiency trade-off.

\begin{figure}[t]
  \centering
  \includegraphics[width=\linewidth]{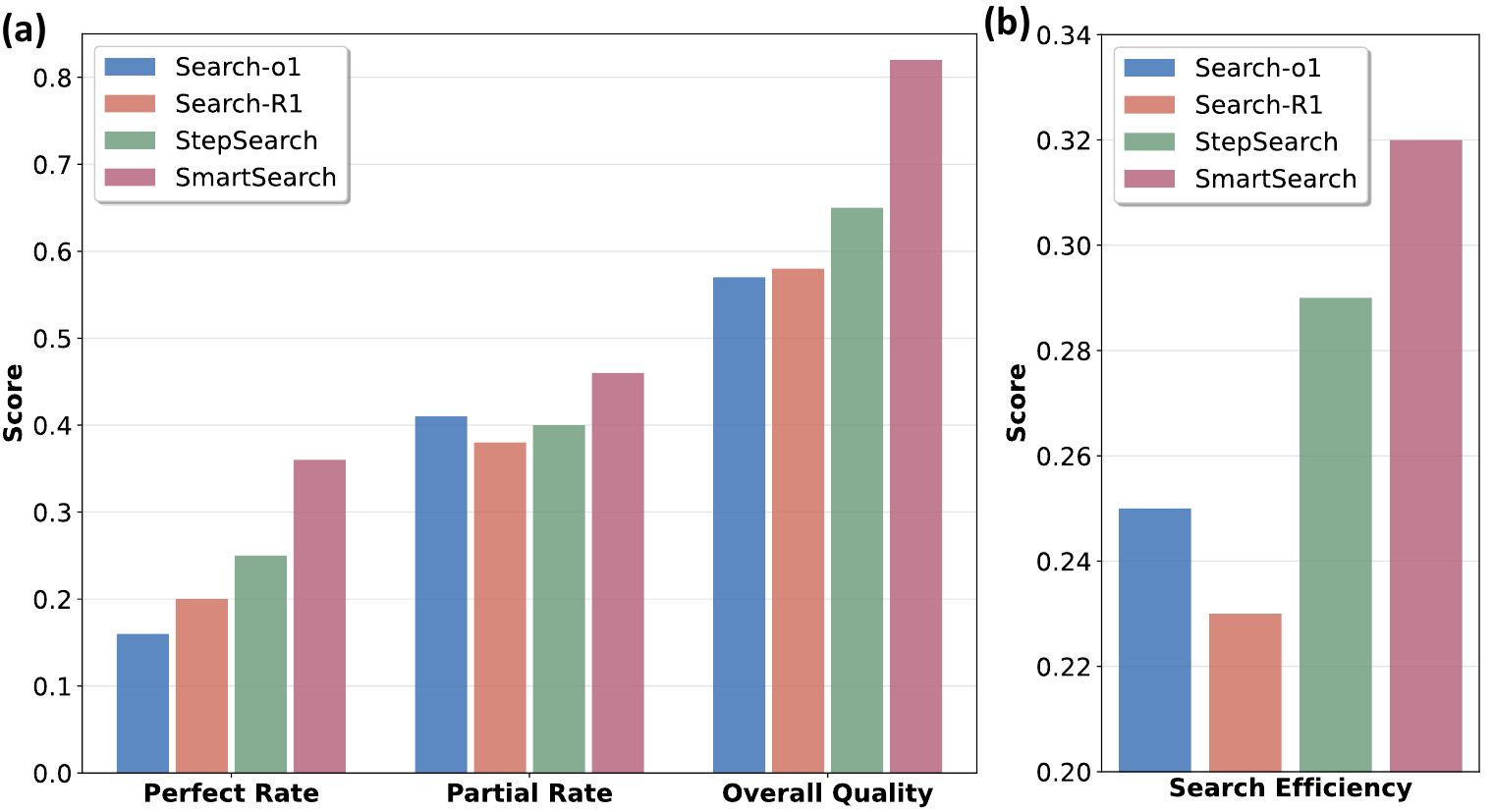}
  \caption{Left: Search query quality comparison. Right: Search efficiency comparison.}
  \label{fig:search efficiency}
\end{figure}

\paragraph{\textbf{Search Query Quality Analysis}} To assess whether \mymodel{} improves the quality of intermediate search queries, we compare the Search Quality metric across various methods. As presented in Figure \ref{fig:search efficiency} (a), \mymodel{} achieves the highest Search Quality, with the highest values for both Perfect Rate and Partial Rate, which contribute to the overall Search Quality metric. This indicates that \mymodel{} effectively enhances the quality of intermediate search queries. Specifically, the search agent reduces ineffective searches while striving to generate perfect trajectories where all queries are of high quality. Even when unable to provide a final correct answer, the agent makes more attempts to generate high-quality queries that edge closer to the correct solution.

\paragraph{\textbf{Search Efficiency Analysis}} The results in previous sections have shown that \mymodel{} outperforms all baselines in accuracy. We now further evaluate whether it also achieves superior search efficiency. To this end, we compare the search efficiency metrics across multiple methods, and as shown in Figure \ref{fig:search efficiency} (b), \mymodel{} outperforms all other methods in this regard. This suggests that by optimizing the quality of intermediate search queries, \mymodel{} generates more precise queries, reducing ineffective or failed search rounds and, as a result, improving overall search efficiency.

\begin{figure}[h]
  \centering
  \includegraphics[width=\linewidth]{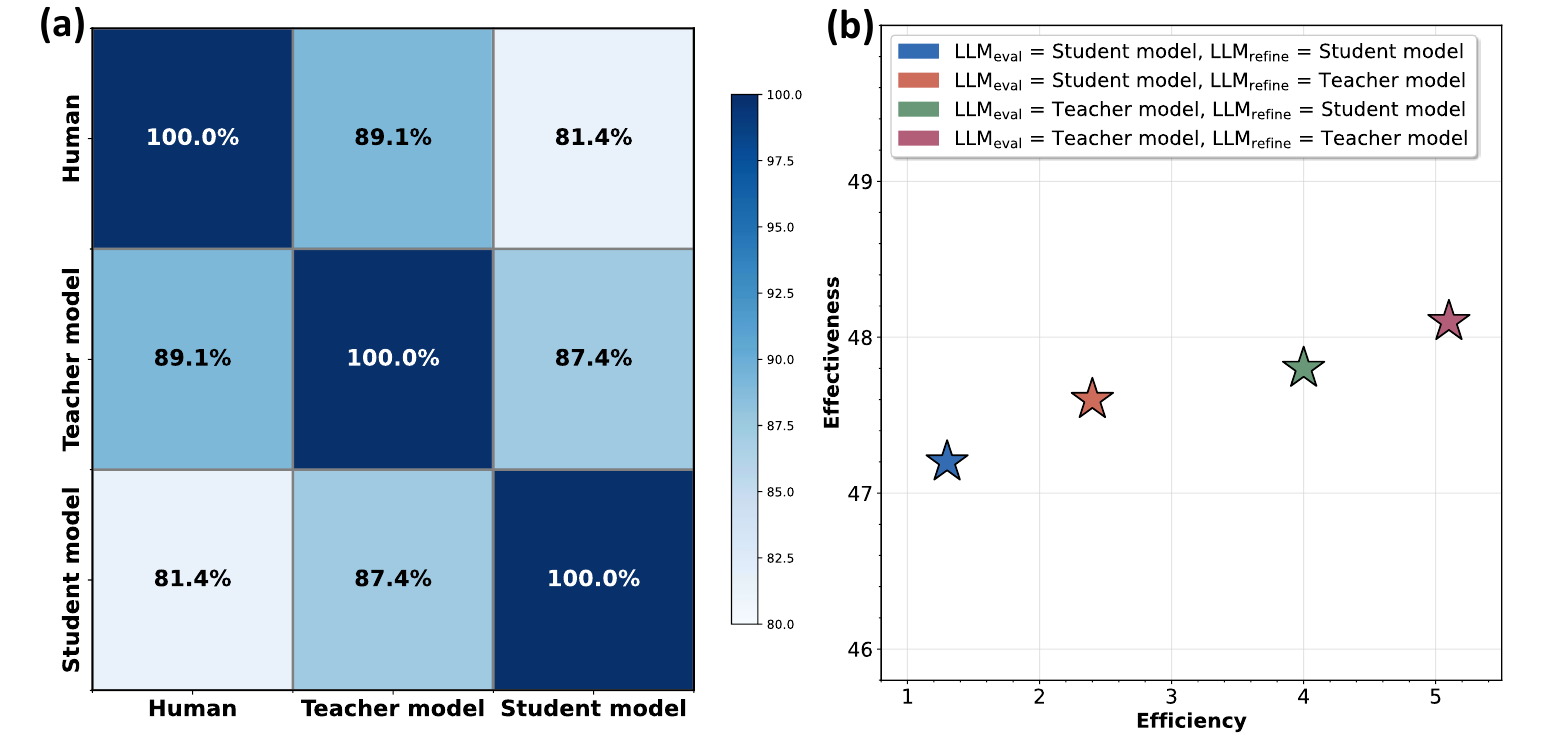}
  \caption{Left: Overlap between the scores assigned to queries by the student model, teacher model, and human annotations. Right: Effectiveness and efficiency tradeoff in \mymodel{}.}
  \label{fig:prm}
\end{figure}

\paragraph{\textbf{Process Reward Model Analysis}}
The process reward model plays a crucial role in our approach by providing fine-grained supervision for the quality of each query and guiding subsequent query refinement. To assess the effectiveness of our process reward model, we randomly choose 100 trajectories. For each search query in these trajectories, we compare the scores annotated by the teacher model, the student model, and human annotators. Figure \ref{fig:prm} (a) illustrates the overlap between the scores assigned to each intermediate search query by these three sources. The results reveal that the teacher model achieves nearly 90\% overlap with human annotations, demonstrating its effectiveness in labeling the training data. After training, the student model achieves over 85\% overlap with the teacher model, indicating the effectiveness of our fine-tuning. Finally, the student model shows over 80\% overlap with human annotations, a result that is entirely acceptable, striking a good balance between scoring accuracy and efficiency.

\paragraph{\textbf{Effectiveness-Efficiency Trade-off}}
To validate the suitability of the lightweight student model as both $\text{LLM}_\text{eval}$ and $\text{LLM}_\text{refine}$, we conduct experiments using a more powerful teacher model in these roles. In this experiment, effectiveness is defined as the average F1 score, while efficiency refers to the average time (s) required to apply the process rewards and query refinement mechanisms to each sample. As shown in Figure \ref{fig:prm} (b), using a more powerful model as both $\text{LLM}_\text{eval}$ and $\text{LLM}_\text{refine}$ indeed improves the agent's effectiveness, underscoring the importance of the process rewards and query refinement mechanisms within our training framework. However, this gain in average F1 score is modest, remaining below 1\%, whereas the time required to process each sample increases by nearly five times. This result demonstrates a clear trade-off between effectiveness and efficiency, and the decision to use the lightweight student model as both $\text{LLM}_\text{eval}$ and $\text{LLM}_\text{refine}$ proves to be a sensible one. This choice achieves an optimal balance between effectiveness and efficiency, ensuring effective query optimization while avoiding excessive computational costs.

\section{Conclusion}
In this work, we introduce \mymodel{}, a framework designed to optimize the quality of intermediate search queries through two key mechanisms: (1) Process rewards, which provide fine-grained supervision for the quality of each query through Dual-Level Assessment. (2) Query refinement, which promotes the optimization of query generation by selectively refining low-quality queries and regenerating subsequent search rounds from these refined points. Building on the foundation of the two mechanisms, we design a three-stage curriculum learning framework that guides the agent through a progression from imitation and alignment to generalization, enabling it to progressively internalize the ability to enhance query quality. Experiments across four challenging benchmarks demonstrate that \mymodel{} consistently surpasses existing baselines, with further quantitative analyses confirming significant gain in both search efficiency and query quality.



\bibliographystyle{ACM-Reference-Format}
\bibliography{main}

\appendix
\section{Prompt Templates}
\label{sec:Prompt Templates}

\begin{tcolorbox}[title = {Prompt for \mymodel{}.}]
You are a helpful assistant that can solve the given question step by step with the help of the wikipedia search tool. Given a question, you need to first think about the reasoning process in the mind and then provide the answer. During thinking, you can invoke the wikipedia search tool to search for fact information about specific topics if needed. The reasoning process and answer are enclosed within <think> </think> and <answer> </answer> tags respectively, and the search query and result are enclosed within <search> </search> and <result> </result> tags respectively. For example, <think> This is the reasoning process. </think> <search> search query here </search> <result> search result here </result> <think> This is the reasoning process. </think> <answer> The final answer is \[ \boxed{\text{answer here}} \] </answer>. In the last part of the answer, the final exact answer is enclosed within \textbackslash boxed\{ \} with latex format.
\end{tcolorbox}

\begin{tcolorbox}[title = {Prompt for scoring.}]
You are a query-evaluation assistant. Your task is to assess the quality of a search agent's query of the current search round according to the user's question, the golden answer and the agent's search process up to the current search round.

If the agent's query intent of the current search round is necessary , and the corresponding query result includes the answer for the query, the score for query should be 1. Otherwise, the score for the query should be 0. The details of the assessment are in the Evaluation Guideline, please read it carefully.

\#\#\# User's question

\{question\}

\#\#\# Golden answer

\{answer\}

\#\#\# Agent's search process up to the current search round

\{context\}

\#\#\# Evaluation Guideline

1. Identify the agent's query intent of the current search round accurately.

2. The query result doesn't need to solve the user's question directly; but it must include the information that address the agent's query intent completely, related entities alone is not enough.

3. The intended entity and the one found in the query result must be exactly the same, otherwise, the score should be 0.

\#\#\# Output Format:

<answer> score for the query </answer>

<explanation> explanation for the score </explanation>

\end{tcolorbox}

\begin{tcolorbox}[title = {Prompt for Refining.}]
You are a query-refine assistant. Your task is to refine a search agent's query of the current search round within <search> </search> according to the user's question, the agent's search process up to the current search round and the issues of the query. The details of the refinement are in the Refine Guideline, please read it carefully.

\#\#\# User's question

\{question\}

\#\#\# Agent's search process up to the current search round

\{context\}

\#\#\# Issues of the query

\{explanation\}

\#\#\# Refine Guideline

1. The refined query is meant to replace the query of the current round, so don't rely on any query result within <result> </result> from the current round when refining the query.

2. If the issues of the query indicate that the query intent is unreasonable, the refined query should serve for a more necessary and actionable query intent.

3. The refined query can be expressed as a complete semantic question or a keyphrase-based query, and you may add or remove information from the original query. All depends on which option best serves the agent's query intent, ensuring that the query result contains the answer to the agent's query intent.

\#\#\# Output format:

<search> refined query </search>

<explanation> explanation for the refined query </explanation>

\end{tcolorbox}

\section{Implementation Details}
\label{sec:Implementation Details}
In the Query Quality Screened Imitation Learning stage, we employ ARPO-14B \cite{dong2025agentic} as the policy model for trajectory sampling. The trajectories obtained through this sampling process are then used to fine-tune the Qwen2.5-3B-Instruct model through SFT, resulting in the SFT model. The training is conducted with a learning rate of 7e-6 over a total of 3 epochs, and the maximum input length during training is set to 16384 tokens. We utilize DeepSpeed ZeRO-3 \cite{rasley2020deepspeed} and FlashAttention2 \cite{dao2024flashattention} to accelerate training, with a total batch size of 64 and applying BF16 precision.

In the Query Generation Alignment stage, we perform DPO training using trajectories generated by the SFT model as positive and negative samples, resulting in the DPO model. This process involves LoRA fine-tuning with a learning rate of 7e-6 trained for 3 epochs, and the maximum input length during training is set to 10000 tokens. As in the previous stage, we leverage DeepSpeed ZeRO-3 and FlashAttention2 for efficient training, with a total batch size of 32 and BF16 precision. 

In the Query Aware Policy Optimization stage, we focus on a curated set of challenging questions that remained unresolved after four sampling trials. Through RL, the DPO model is further optimized to produce the final \mymodel{} model. The training is conducted with a learning rate of 1e-6, where each sample undergoes 8 rollouts to explore different trajectories. The total training batch size is 64, with a PPO mini-batch size of 16. We set the maximum output length to 8192 tokens and limit the number of tool calls to 5 during each rollout.

During the Inference stage, we set the maximum output length to 16384 tokens and limit the number of tool calls to 10. 

\section{Case Study}
To better demonstrate the effectiveness of \mymodel{}, as well as the process reward and query refinement mechanisms, we conducted a case study.

As shown in Table \ref{tab:Example of Output}, when handling a user's question, \mymodel{} first comprehends the question and performs preliminary planning to form an accurate search intent. Based on this intent, the model formulate a precise search query and successfully retrieves the desired information. It is also noteworthy that our model effectively utilizes its internal knowledge, generating only necessary search queries, which further enhances the search efficiency of \mymodel{}.

Table \ref{tab:Example of Process Reward} illustrates how process reward mechanism provides both numerical scores and textual explanations for each search query within the model’s output. The scores accurately reflect the quality of the search queries, while the explanations offer detailed feedback. For low‑quality queries, the explanations clearly identify the reasons for their subpar performance, providing crucial guidance for subsequent refinement.

As presented in Table \ref{tab:Example of Query Refinement}, query refinement mechanism refines the low‑quality queries based on the explanations provided by the process reward and regenerates the subsequent steps from the refined query. The initial query fails to retrieve the expected information, causing the entire trajectory to deviate from the correct path. In contrast, the refined query successfully retrieves the desired information and leads to the correct answer. The comparison between the two trajectories helps the model better concentrate its optimization on search queries, highlighting the effectiveness of this mechanism.
\begin{table*}[!t]
    \centering
    \caption{Example of \mymodel{} output, including the question, golden answer, and model output.}
    \fontsize{9pt}{11pt}\selectfont
    \renewcommand{\arraystretch}{0.8} 
    \begin{tabular}{p{0.95\linewidth}}
    \toprule
        \rowcolor{gray!20}
        \textbf{Example of \mymodel{} Output} \\
    \midrule
        \textbf{Question:}
        
        Douglas D. Scott is an American archaeologist most notable for his work at the site of a battle that occurred in what year?\\\\
        \textbf{Golden Answer:}
        
        1876 \\\\
        \textbf{Model Output:}
    
        $<$think$>$ Okay, so I need to figure out the year of the battle that Douglas D. Scott worked on. Let me start by recalling what I know about Douglas D. Scott. He's an American archaeologist, so maybe he's known for excavating a significant site. The question mentions a battle site, so I should look for his notable archaeological sites related to battles. First, I'll try to search for Douglas D. Scott's main archaeological sites. Maybe he's famous for a specific battle site. Let me start with a general search query. $<$/think$>$\\\\
        $<$search$>$ Douglas D. Scott notable archaeological sites $<$/search$>$\\\\
        $<$result$>$ result: ``Douglas D. Scott'' by Ronald K. Wetherington and Frances Levine, pp 134–152, University of Oklahoma Press, Norman. Douglas D. Scott Douglas D. Scott is an American archaeologist most notable for his work at the Little Bighorn in the mid-1980s. Working with Richard Fox, Melissa Connor, Doug Harmon, and staff and volunteers from the National Park Service, Scott worked to sketch out a field methodology that has enabled archaeologists to systematically investigate battlefields. This work is internationally recognized as constituting a great step forward in our ability to interpret battlefields archaeologically, regardless of the extent of the historical record. At the Little Bighorn, the ... $<$/result$>$\\\\
        $<$think$>$ Hmm, looking at the search results, the first entry mentions Douglas D. Scott's work at the Little Bighorn in the mid-1980s. The other entries don't seem to mention a battle site. The Little Bighorn is a well-known battle, so that's probably the one. The question asks for the year of the battle. The Little Bighorn was a major battle during the American Indian Wars, specifically between the Lakota, Northern Cheyenne, and Arapaho tribes against the U.S. Army. The Battle of the Little Bighorn took place on June 25, 1876. Let me confirm the exact year. The search result says `mid-1980s' for his work at Little Bighorn. Since the battle itself was in 1876, the year would be 1876. I don't see any conflicting information here. The other sites mentioned are not battles but other archaeological sites. So the answer should be 1876. $<$/think$>$\\\\
        $<$answer$>$ The final answer is \boxed{1876}. $<$/answer$>$ \\
    \bottomrule
    \end{tabular}
    \label{tab:Example of Output}
\end{table*}

\begin{table*}[!t]
    \centering
    \caption{Example of process reward, including the question, golden answer, model output, and process reward for each query within the model output.}
    \fontsize{9pt}{11pt}\selectfont
    \renewcommand{\arraystretch}{0.8} 
    \begin{tabular}{p{0.95\linewidth}}
    \toprule
        \rowcolor{gray!20}
        \textbf{Example of Process Reward} \\
    \midrule
        \textbf{Question:}
        
        An Annapolis Story stars which American stage, film, and television actor born on February 15, 1914?\\\\
        \textbf{Golden Answer:}
        
        Kevin McCarthy \\\\
        \textbf{Model Output:}
    
        $<$think$>$ To solve this question, I need to find out which American stage, film, and television actor born on February 15, 1914, stars in the movie ``An Annapolis Story''. Step 1: Identify the actor who stars in ``An Annapolis Story''. Step 2: Determine which of that actor's birthdate matches February 15, 1914. Let's start with Step 1: Step 1: Identify the actor who stars in ``An Annapolis Story''. I'll use a query to search for information about the stars of ``An Annapolis Story'': $<$/think$>$\\\
        $<$search$>$ stars in ``An Annapolis Story'' $<$/search$>$\\\\
        $<$result$>$ result:``An Annapolis Story'' An Annapolis Story An Annapolis Story (alternative titles: The Blue and Gold and Navy Air Patrol) is a 1955 American drama film directed by Don Siegel and starring John Derek, Diana Lynn and Kevin McCarthy.A product of the newly formed Allied Artists company, ``An Annapolis Story'', despite having the ``Siegel Touch'', suffered from its low budget. Brothers Tony (John Derek) and Jim Scott (Kevin McCarthy) enroll as midshipmen at the United States Naval Academy in Annapolis. Jim, the older one, looks after the more impulsive Tony and helps him pass a difficult test so he can play football in the ... $<$/result$>$\\\\
        $<$think$>$ From the search results, I can see that ``An Annapolis Story'' is a 1955 American drama film starring John Derek, Diana Lynn, and Kevin McCarthy. Step 2: Determine which of that actor's birthdate matches February 15, 1914. I'll use a query to search for information about the birthdate of Kevin McCarthy: $<$/think$>$\\\\
        $<$search$>$ birthdate of Kevin McCarthy $<$/search$>$ \\\\
        $<$result$>$ result: ``Kevin McCarthy (California politician)'' Kevin McCarthy (California politician) Kevin Owen McCarthy (born January 26, 1965) is an American politician serving as the House Majority Leader since 2014 and U.S. Representative for California's 23rd congressional district since 2007. The 23rd district, numbered as the 22nd district from 2007 to 2013, is based in Bakersfield and includes large sections of Kern County and Tulare County as well as part of the Quartz Hill neighborhood in northwest Los Angeles County. A member of the Republican Party, he was formerly chairman of the California Young Republicans and the Young Republican National Federation. McCarthy worked as district director for ... $<$/result$>$\\\\
        $<$think$>$ From the search results, I can see that Kevin McCarthy was born on January 26, 1965. Therefore, the American stage, film, and television actor born on February 15, 1914, who stars in ``An Annapolis Story'' is not Kevin McCarthy. However, the question might be referring to another actor. ... $<$/think$>$\\\\
        \textbf{Process Reward:}

        Score1: 1
        
        Explanation1: The search intent is necessary, and the search results included the cast list for the movie An Annapolis Story.

        Score2: 0
        
        Explanation2: The search intent is necessary, but the search results did not include the birth date of actor Kevin McCarthy. Instead, they contained information about politician Kevin McCarthy. Therefore, the score is 0.
        \\
    \bottomrule
    \end{tabular}
    \label{tab:Example of Process Reward}
\end{table*}

\begin{table*}[!t]
    \centering
    \caption{Example of query refinement, including the question, golden answer, model output, refined query, and the regenerated subsequent steps.}
    \fontsize{9pt}{11pt}\selectfont
    \renewcommand{\arraystretch}{0.8} 
    \begin{tabular}{p{0.95\linewidth}}
    \toprule
        \rowcolor{gray!20}
        \textbf{Example of Query Refinement} \\
    \midrule
        \textbf{Question:}
        
        An Annapolis Story stars which American stage, film, and television actor born on February 15, 1914?\\\\
        \textbf{Golden Answer:}
        
        Kevin McCarthy \\\\
        \textbf{Original Model Output:}
    
        $<$think$>$ To solve this question, I need to find out which American stage, film, and television actor born on February 15, 1914, stars in the movie ``An Annapolis Story''. Step 1: Identify the actor who stars in ``An Annapolis Story''. Step 2: Determine which of that actor's birthdate matches February 15, 1914. Let's start with Step 1: Step 1: Identify the actor who stars in ``An Annapolis Story''. I'll use a query to search for information about the stars of ``An Annapolis Story'': $<$/think$>$\\\
        $<$search$>$ stars in ``An Annapolis Story'' $<$/search$>$\\\\
        $<$result$>$ result: ``An Annapolis Story'' An Annapolis Story An Annapolis Story (alternative titles: The Blue and Gold and Navy Air Patrol) is a 1955 American drama film directed by Don Siegel and starring John Derek, Diana Lynn and Kevin McCarthy.A product of the newly formed Allied Artists company, ``An Annapolis Story'', despite having the ``Siegel Touch'', suffered from its low budget. Brothers Tony (John Derek) and Jim Scott (Kevin McCarthy) enroll as midshipmen at the United States Naval Academy in Annapolis. Jim, the older one, looks after the more impulsive Tony and helps him pass a difficult test so he can play football in the ... $<$/result$>$\\\\
        $<$think$>$ From the search results, I can see that ``An Annapolis Story'' is a 1955 American drama film starring John Derek, Diana Lynn, and Kevin McCarthy. Step 2: Determine which of that actor's birthdate matches February 15, 1914. I'll use a query to search for information about the birthdate of Kevin McCarthy: $<$/think$>$\\\\
        $<$search$>$ birthdate of Kevin McCarthy $<$/search$>$ \\\\
        $<$result$>$ result: ``Kevin McCarthy (California politician)'' Kevin McCarthy (California politician) Kevin Owen McCarthy (born January 26, 1965) is an American politician serving as the House Majority Leader since 2014 and U.S. Representative for California's 23rd congressional district since 2007. The 23rd district, numbered as the 22nd district from 2007 to 2013, is based in Bakersfield and includes large sections of Kern County and Tulare County as well as part of the Quartz Hill neighborhood in northwest Los Angeles County. A member of the Republican Party, he was formerly chairman of the California Young Republicans and the Young Republican National Federation. McCarthy worked as district director for ... $<$/result$>$\\\\
        $<$think$>$ From the search results, I can see that Kevin McCarthy was born on January 26, 1965. Therefore, the American stage, film, and television actor born on February 15, 1914, who stars in ``An Annapolis Story'' is not Kevin McCarthy. However, the question might be referring to another actor. ... $<$/think$>$\\\\
        \textbf{Query Refinement}
        
        $<$search$>$ birthdate of Kevin McCarthy $<$/search$>$ \textbf{$\to$}
        $<$search$>$ birthdate of Actor Kevin McCarthy $<$/search$>$\\\\
        \textbf{Regenerate Subsequent Steps}

        $<$search$>$ birthdate of Actor Kevin McCarthy $<$/search$>$\\\\
        $<$result$>$ result: ``Kevin McCarthy (actor)'' Kevin McCarthy (actor) Kevin McCarthy (February 15, 1914 – September 11, 2010) was an American actor who gave over 200 television and film performances. He is best remembered for portraying the male lead in the horror science fiction film ``Invasion of the Body Snatchers'' (1956). Following several television guest roles, McCarthy gave his first credited film performance in ``Death of a Salesman'' (1951), portraying Biff Loman to Fredric March's Willy Loman. The role earned him a Golden Globe Award and a nomination for the Academy Award for Best Supporting Actor. McCarthy was born in Seattle, Washington, the son of Roy ... $<$/result$>$\\\\
        $<$think$>$ The search results confirm that Kevin McCarthy was an American actor born on February 15, 1914. He starred in ``An Annapolis Story'' (1955) and had a long career in film, television, and stage. The question asks for an actor who starred in ``An Annapolis Story'' and was born on February 15, 1914. Kevin McCarthy fits all these criteria. $<$/think$>$\\\\
        <answer> \boxed{Kevin\ McCarthy} </answer>\\
    \bottomrule
    \end{tabular}
    \label{tab:Example of Query Refinement}
\end{table*}

\end{document}